\documentclass[11pt,letterpaper]{article}
\usepackage{emnlp2017}
\usepackage{times}
\usepackage{amsfonts}
\usepackage{latexsym}
\usepackage{url}
\usepackage{makecell}
\usepackage{hhline}
\usepackage{multirow, rotating}
\usepackage{bbding}
\usepackage{amsfonts, amssymb, amsmath}
\usepackage{algorithm, algorithmic, amsthm}
\usepackage{graphicx}
\usepackage{color}
\usepackage{float}
\usepackage{paralist}

\usepackage{subfigure}
\usepackage{todonotes}
\newcommand{\textbox}[2]{{\color{#1}\fbox{\normalcolor#2}}}

\emnlpfinalcopy

\def\emnlppaperid{1176}

\title{Composite Task-Completion Dialogue Policy Learning via \\ Hierarchical Deep Reinforcement Learning}

\author{Baolin Peng$^{\star}$\quad Xiujun Li$^{\dagger}$\quad Lihong Li$^{\dagger}$\quad Jianfeng Gao$^{\dagger}$\quad \\ 
\textbf{Asli Celikyilmaz$^\dagger$}\quad \textbf{Sungjin Lee$^\dagger$}\quad \textbf{Kam-Fai Wong$^{\star}$}\\
  $^{\dagger}$Microsoft Research, Redmond, WA, USA\\
  $^{\star}$The Chinese University of Hong Kong, Hong Kong \\
  {\tt \{blpeng, kfwong\}@se.cuhk.edu.hk} \\
  {\tt \{xiul,lihongli,jfgao,aslicel,sule\}@microsoft.com}
}

\date{}

\begin{document}
	
\maketitle
	
\begin{abstract}
Building a dialogue agent to fulfill complex tasks, such as travel planning, is challenging because the agent has to learn to \emph{collectively} complete multiple subtasks. For example, the agent needs to reserve a hotel and book a flight so that there leaves enough time for commute between arrival and hotel check-in. This paper addresses this challenge by formulating the task in the mathematical framework of \emph{options} over Markov Decision Processes (MDPs), and proposing a hierarchical deep reinforcement learning approach to learning a dialogue manager that operates at different temporal scales. The dialogue manager consists of: (1) a top-level dialogue policy that selects among subtasks or options, (2) a low-level dialogue policy that selects primitive actions to complete the subtask given by the top-level policy, and (3) a global state tracker that helps ensure all cross-subtask constraints be satisfied. Experiments on a travel planning task with simulated and real users show that our approach leads to significant improvements over three baselines, two based on handcrafted rules and the other based on flat deep reinforcement learning.

\end{abstract}

\section{Introduction}
There is a growing demand for intelligent personal assistants, mainly in the form of dialogue agents, that can help users accomplish tasks ranging from meeting scheduling to vacation planning. However, most of the popular agents in today's market, such as Amazon Echo, Apple Siri, Google Home and Microsoft Cortana, can only handle very simple tasks, such as reporting weather and requesting songs. Building a dialogue agent to fulfill complex tasks 
remains one of the most fundamental challenges for the NLP community and AI in general.


In this paper, we consider an important type of complex tasks, termed \emph{composite task}, which consists of a set of subtasks that need to be fulfilled collectively. For example, in order to make a travel plan, we need to book air tickets, reserve a hotel, rent a car, etc. in a collective way so as to satisfy a set of cross-subtask constraints, which we call \emph{slot constraints}. Examples of slot constraints for travel planning are: hotel check-in time should be later than the flight's arrival time, hotel check-out time may be earlier than the return flight depart time, the number of flight tickets equals to that of hotel check-in people, and so on.


It is common to learn a task-completion dialogue agent using reinforcement learning (RL); see \citet{su2016continuously,cuayahuitl2017simpleds,williams2017hybrid,Dhingra17EndToEnd} and \citet{li2017end} for a few recent examples. Compared to these dialogue agents developed for individual domains, the composite task presents additional challenges to commonly used, \emph{flat} RL approaches such as DQN~\cite{DBLP:journals/nature/MnihKSRVBGRFOPB15}. The first challenge is reward sparsity. 
Dialogue policy learning for composite tasks requires exploration in a much larger state-action space, and it often takes many more conversation turns between user and agent to fulfill a task, leading to a much longer trajectory. Thus, the reward signals (usually provided by users at the end of a conversation) are delayed and sparse. As we will show in this paper, typical flat RL methods such as DQN with naive $\epsilon$-greedy exploration is rather inefficient. The second challenge is to satisfy slot constraints across subtasks. This requirement makes most of the existing methods of learning \emph{multi-domain dialogue} agents~\cite{cuayahuitl2009hierarchical,DBLP:conf/asru/GasicMSVWY15} inapplicable: these methods train a collection of policies, one for each domain, and there is no cross-domain constraints required to successfully complete a dialogue. 
The third challenge is improved user experience: we find in our experiments that a flat RL agent tends to switch between different subtasks frequently when conversing with users. Such incoherent conversations lead to poor user experience, and are one of the main reasons that cause a dialogue session to fail.


In this paper, we address the above mentioned challenges by formulating the task using the mathematical framework of \emph{options over MDPs}~\cite{DBLP:journals/ai/SuttonPS99}, and proposing a method that combines deep reinforcement learning and hierarchical task decomposition to train a composite task-completion dialogue agent. At the heart of the agent is a dialogue manager, which consists of (1) a top-level dialogue policy that selects subtasks (options), (2) a low-level dialogue policy that selects primitive actions to complete a given subtask, and (3) a global state tracker that helps ensure all cross-subtask constraints be satisfied.

Conceptually, our approach exploits the structural information of composite tasks for efficient exploration. Specifically, in order to mitigate the reward sparsity issue, we equip our agent with an evaluation module (internal critic) that gives intrinsic reward signals, indicating how likely a particular subtask is completed based on its current state generated by the global state tracker. Such intrinsic rewards can be viewed as heuristics that encourage the agent to focus on solving a subtask 
before moving on to another subtask. 
Our experiments show that such intrinsic rewards can be used inside a hierarchical RL agent to make exploration more efficient, yielding a significantly reduced state-action space for decision making. Furthermore, it leads to a better user experience, as the resulting conversations switch between subtasks less frequently.

To the best of our knowledge, this is the first work that strives to develop a composite task-completion dialogue agent. Our main contributions are three-fold:

\begin{itemize}
\item We formulate the problem in the mathematical framework of options over MDPs.
\item We propose a hierarchical deep reinforcement learning approach to efficiently learning the dialogue manager that operates at different temporal scales. 
\item We validate the effectiveness of the proposed approach in a travel planning task on simulated as well as real users. 
\end{itemize}

\section{Related Work}
Task-completion dialogue systems have attracted numerous research efforts. Reinforcement learning algorithms hold the promise for dialogue policy optimization over time with experience~\cite{DBLP:conf/icassp/SchefflerY00,DBLP:journals/taslp/LevinPE00,DBLP:journals/pieee/YoungGTW13,williams2017hybrid}. Recent advances in deep learning have inspired many deep reinforcement learning based dialogue systems that eliminate the need for feature engineering~\cite{su2016continuously,cuayahuitl2017simpleds,williams2017hybrid,Dhingra17EndToEnd,li2017end}.

All the work above focuses on single-domain problems. Extensions to composite-domain dialogue problems are non-trivial due to several reasons: the state and action spaces are much larger, the trajectories are much longer, and in turn reward signals are much more sparse.
All these challenges 
can be addressed by hierarchical reinforcement learning~\cite{DBLP:journals/ai/SuttonPS99,DBLP:conf/icml/SuttonPS98,DBLP:conf/aaai/Singh92,DBLP:journals/jair/Dietterich00,DBLP:journals/deds/BartoM03}, which decomposes a complicated task into simpler subtasks, possibly in a recursive way.  Different frameworks have been proposed, such as Hierarchies of Machines~\cite{DBLP:conf/nips/ParrR97} and MAXQ decomposition~\cite{DBLP:journals/jair/Dietterich00}.  In this paper, we choose the \emph{options framework} for its conceptual simplicity and generality~\cite{DBLP:conf/icml/SuttonPS98}; more details are found in the next section.  Our work is also motivated by hierarchical-DQN~\cite{DBLP:conf/nips/KulkarniNST16} which integrates hierarchical value functions to operate at different temporal scales. The model achieved superior performance on a complicated ATARI game ``Montezuma's Revenge'' with a hierarchical structure.

A related but different extension to single-domain dialogues is multi-domain dialogues, where each domain is handled by a separate agent~\cite{DBLP:conf/sigdial/Lison11,DBLP:conf/icassp/GasicKTY15,DBLP:conf/asru/GasicMSVWY15,cuayahuitl2016deep}. In contrast to composite-domain dialogues studied in this paper, a conversation in a multi-domain dialogue normally involves one domain, so completion of a task does \textit{not} require solving sub-tasks in different domains. Consequently, work on multi-domain dialogues focuses on different technical challenges such as transfer learning across different domains~\cite{DBLP:conf/icassp/GasicKTY15} and domain selection~\cite{cuayahuitl2016deep}.

\section{Dialogue Policy Learning}
Our composite task-completion dialogue agent consists of four components: (1) an LSTM-based language understanding module~\cite{hakkani2016multi, DBLP:conf/slt/YaoPZYZS14} for identifying user intents and extracting associated slots; (2) a state tracker for tracking the dialogue state; (3) a dialogue policy which selects the next action based on the current state; and (4) a model-based natural language generator~\cite{DBLP:conf/emnlp/WenGMSVY15} for converting agent actions to natural language responses. Typically, a dialogue manager contains a state tracker and a dialogue policy. In our implementation, we use a \emph{global} state tracker to maintain the dialogue state by accumulating information across all subtasks, thus helping ensure all inter-subtask constraints be satisfied. In the rest of this section, we will describe the dialogue policy in details.

\begin{figure}[htb]
\centering
\includegraphics[width=0.7\linewidth]{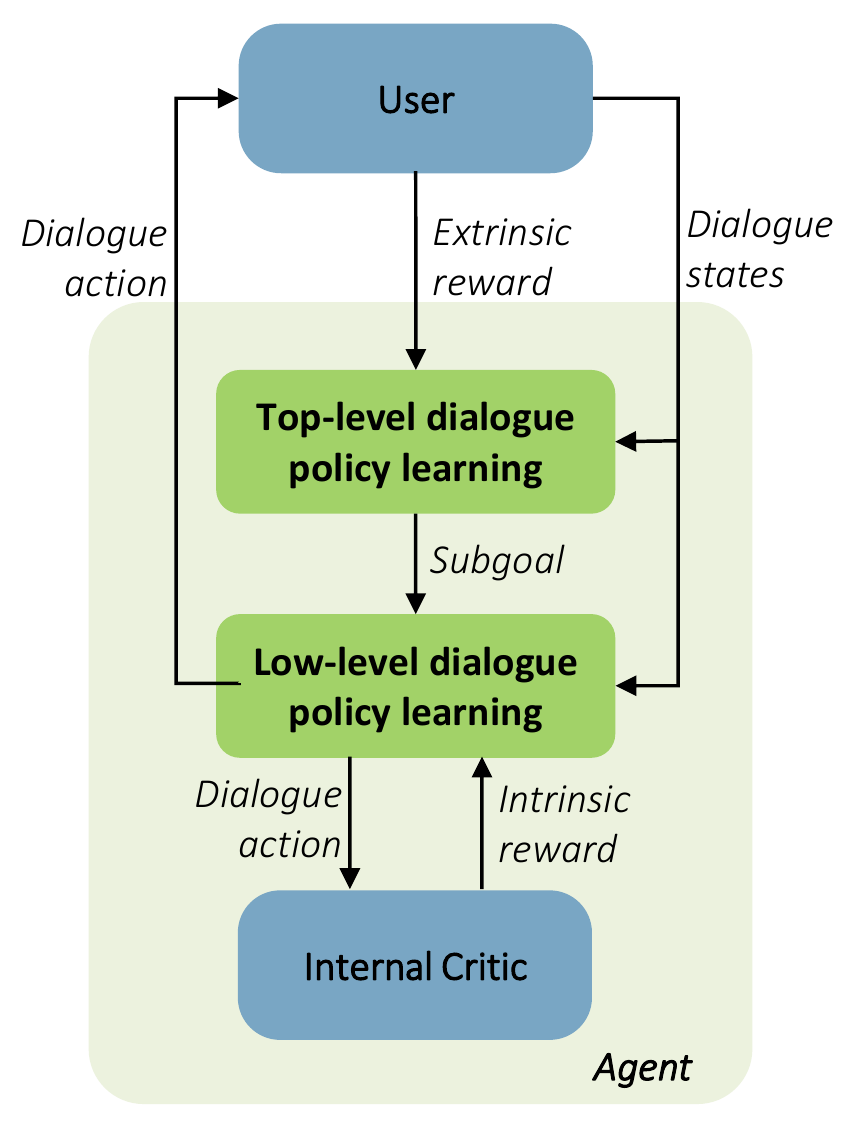}
\vspace{-3mm}
\caption{Overview of a composite task-completion dialogue agent.}
\label{fig:critic}
\end{figure}

\subsection{Options over MDPs}

Consider the following process of completing a composite task (e.g., travel planning). An agent first selects a subtask (e.g., book-flight-ticket), then takes a sequence of actions to gather related information (e.g., departure time, number of tickets, destination, etc.) until all users' requirements are met and the subtask is completed, and finally chooses the next subtask (e.g., reserve-hotel) to complete. The composite task is fulfilled after all its subtasks are completed collectively. The above process has a natural hierarchy: a top-level process selects which subtasks to complete, and a low-level process chooses primitive actions to complete the selected subtask. Such hierarchical decision making processes can be formulated in the \emph{options} framework~\cite{DBLP:journals/ai/SuttonPS99}, where options generalize primitive actions to higher-level actions.  Different from the traditional MDP setting where an agent can only choose a primitive action at each time step, with options the agent can choose a ``multi-step" action which for example could be a sequence of primitive actions for completing a subtask. As pointed out by \newcite{DBLP:journals/ai/SuttonPS99}, options are closely related to actions in a family of decision problems known as semi-Markov decision processes.

Following~\newcite{DBLP:journals/ai/SuttonPS99}, an option consists of three components: a set of states where the option can be initiated, an intra-option policy that selects primitive actions while the option is in control, and a termination condition that specifies when the option is completed.  For a composite task such as travel planning, subtasks like \textit{book-flight-ticket} and \textit{reserve-hotel} can be modeled as options. Consider, for example, the option \textit{book-flight-ticket}: its initiation state set contains states in which the tickets have not been issued or the destination of the trip is long away enough that a flight is needed; it has an intra-option policy for requesting or confirming information regarding departure date and the number of seats, etc.; it also has a termination condition for confirming that all information is gathered and correct so that it is ready to issue the tickets.

\begin{figure}[htb]
\centering
\includegraphics[width=1\linewidth]{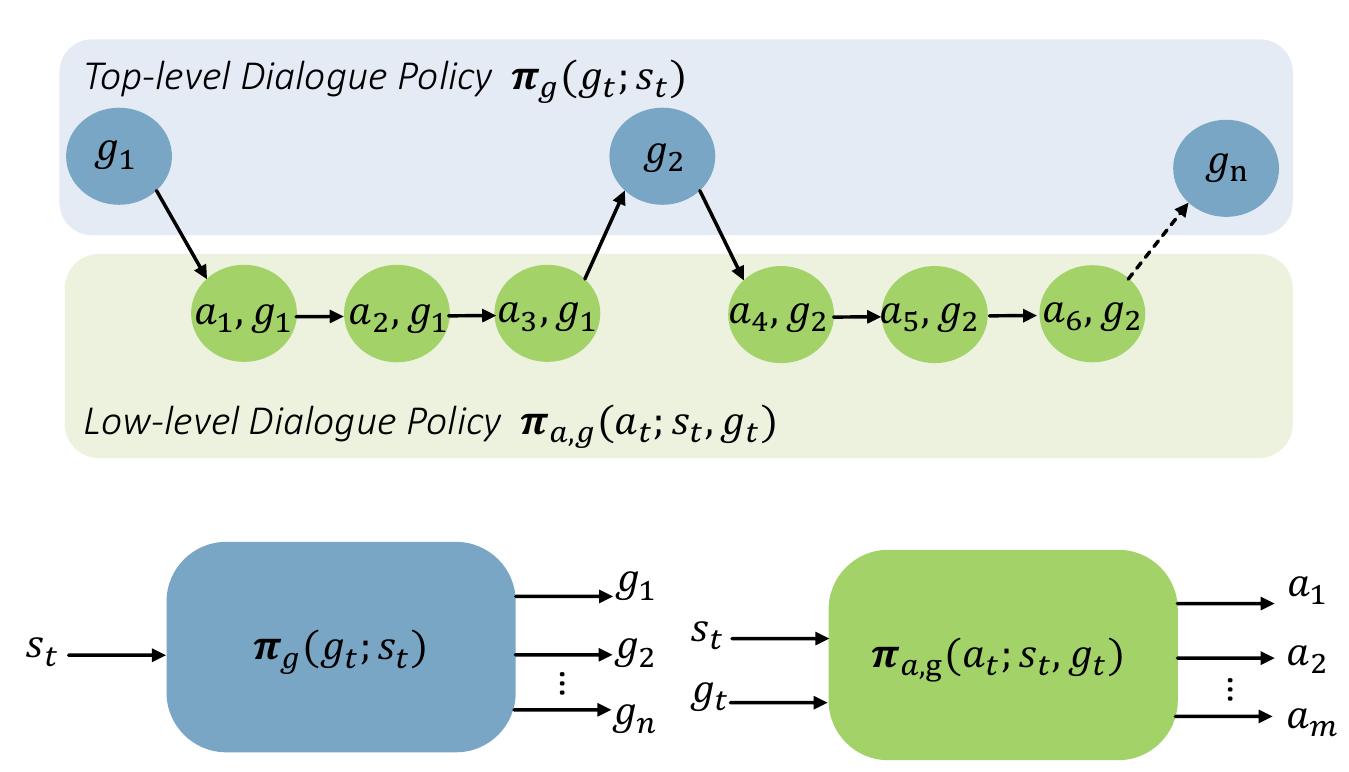}
\vspace{-9mm}
\caption{Illustration of a two-level hierarchical dialogue policy learner.}
\label{fig:hdqn}
\end{figure}

\subsection{Hierarchical Policy Learning}
The intra-option is a conventional policy over primitive actions, we can consider an inter-option policy over sequences of options in much the same way as we consider the intra-option policy over sequences of actions. 
We propose a method that combines deep reinforcement learning and hierarchical value functions to learn a composite task-completion dialogue agent as shown in Figure~\ref{fig:critic}.
It is a two-level hierarchical reinforcement learning agent that consists of a top-level dialogue policy $\pi_g$ and a low-level dialogue policy $\pi_{a,g}$, as shown in Figure~\ref{fig:hdqn}. The top-level policy $\pi_g$ perceives state $s$ from the environment and selects a subtask $g \in \mathcal{G}$, where $\mathcal{G}$ is the set of all possible subtasks.  The low-level policy $\pi_{a,g}$ is shared by all options.  It takes as input a state $s$ and a subtask $g$, and outputs a primitive action $a \in \mathcal{A}$, where $\mathcal{A}$ is the set of primitive actions of all subtasks. 
The subtask $g$ remains a constant input to $\pi_{a,g}$, until a terminal state is reached to terminate $g$. The internal critic in the dialogue manager provides intrinsic reward $r^i_t(g_t)$, indicating whether the subtask $g_t$ at hand has been solved; this signal is used to optimize $\pi_{a,g}$.
Note that the state $s$ contains global information, in that it keeps track of information for all subtasks. 

Naturally, we aim to optimize the low-level policy $\pi_{a,g}$ so that it maximizes the following cumulative intrinsic reward at every step $t$:
%
\begin{eqnarray*}
\max_{\pi_{a,g}} \mathbb{E}\Big[\sum_{k \ge 0} \gamma^{k}r_{t+k}^i \Big| s_t=s, g_t=g,a_{t+k}=\pi_{a,g}(s_{t+k})\Big]\,,
\end{eqnarray*}
where $r_{t+k}^i$ denotes the reward provided by the internal critic at step $t+k$.
Similarly, we want the top-level policy $\pi_g$ to optimize the cumulative extrinsic reward at every step $t$: 
\begin{eqnarray*}
\max_{\pi_g} \mathbb{E}\Big[\sum_{k \ge 0} \gamma^{k}r_{t+k}^e \Big| s_t=s, 
a_{t+k}=\pi_g(s_{t+k})\Big]\,,
\end{eqnarray*}
where $r^e_{t+k}$ is the reward received from the environment at step $t+k$ when a new subtask starts.

Both the top-level and low-level policies can be learned with deep Q-learning methods, like DQN. Specifically, the top-level dialogue policy estimates the optimal Q-function that satisfies the following:
\begin{eqnarray}
\lefteqn{Q_{1}^*(s,g) = \mathbb{E} \Big[ \sum_{k=0}^{N-1} \gamma^k r^e_{t+k} +} \nonumber \\
&\gamma^N \cdot \max_{g'} Q_{1}^*(s_{t+N},g')|s_t=s, g_t=g\Big], \label{eqn:top-level-bellman}
\end{eqnarray}
where $N$ is the number of steps that the low-level dialogue policy (intra-option policy) needs to accomplish the subtask. $g'$ is the agent's next subtask in state $s_{t+N}$. Similarly, the low-level dialogue policy estimates the Q-function that satisfies the following:
\begin{eqnarray*}
\lefteqn{Q_{2}^*(s,a, g) = \mathbb{E} \Big[ r^i_t +} \\
&&\gamma \cdot \max_{a_{t+1}} Q_{2}^*(s_{t+1},a_{t+1}, g)|s_t=s, g_t=g\Big]\,.
\end{eqnarray*}
Both $Q_1^*(s, g)$ and $Q_2^*(s, a, g)$ are represented by neural networks, $Q_1(s, g; \theta_1)$ and $Q_2(s, a, g;\theta_2)$, parameterized by $\theta_1$ and $\theta_2$, respectively.

The top-level dialogue policy tries to minimize the following loss function at each iteration $i$:
\begin{align*}
\mathcal{L}_1(\theta_{1,i}) &= \mathbb{E}_{(s,g,r^e,s')\sim \mathcal{D}_1}[(y_i - Q_1(s,g;\theta_{1,i}))^2]\, \\
y_i &= r^e +\gamma^N \max_{g'}Q_1(s',g',\theta_{1,i-1})\,,
\end{align*}
where, as in Equation~(\ref{eqn:top-level-bellman}), $r^e=\sum_{k=0}^{N-1} \gamma^k r^e_{t+k}$ is the discounted sum of reward collected when subgoal $g$ is being completed, and $N$ is the number of steps $g$ is completed.

The low-level dialogue policy minimizes the following loss at each iteration $i$ using:
\begin{eqnarray*}
\mathcal{L}_2(\theta_{2,i}) &=& \mathbb{E}_{(s,g,a,r^i,s')\sim  \mathcal{D}_2}[(y_i - \\
&& \ \ \ \ \ \ \ \ \ \ \ \ \ \ \ \  \ \ \  Q_2(s,g,  a;\theta_{2,i}))^2] \\
 y_i &=& r^i +\gamma \max_{a'}Q_2(s',g, a',\theta_{2,i-1})\,.
\end{eqnarray*}
We use SGD to minimize the above loss functions. The gradient for the top-level dialogue policy yields:
\begin{equation}
\label{eq:g1}
\begin{split}
\nabla_{\theta_{1,i}}L_1(\theta_{1,i}) &= \mathbb{E}_{(s,g,r^e,s')\sim \mathcal{D}_1} [(r^e + \\\gamma^N \max_{g'} & Q_2(s',g',\theta_{1,i-1}) - Q_1(s,g,\theta_{1,i})) \\ & \nabla_{\theta_{1,i}}Q_1(s,g,\theta_{1,i})]
\end{split}
\end{equation}
The gradient for the low-level dialogue policy yields:
\begin{equation}
\label{eq:g2}
\begin{split}
\begin{split}
\nabla_{\theta_{2,i}}L_2(\theta_{2,i}) &= \mathbb{E}_{(s,g,a,r^i,s')\sim \mathcal{D}_2} [(r^i + \\
\gamma \max_{a'}  Q_2&(s',g,a',\theta_{2,i-1})  - Q_2(s,g,a,\theta_{2,i})) \\
  & \nabla_{\theta_{2,i}}Q_2(s,g,a,\theta_{2,i})]
\end{split}
\end{split}
\end{equation}
Following previous studies, we apply two most commonly used performance boosting methods: target networks and experience replay. Experience replay tuples $(s,g,r^e,s')$ and $(s,g,a,r^i,s')$, are sampled from the experience replay buffers $\mathcal{D}_1$ and $\mathcal{D}_2$ respectively. A detailed summary of the learning algorithm for the hierarchical dialogue policy is provided in Appendix B.

\section{Experiments and Results}
To evaluate the proposed method, we conduct experiments on the composite task-completion dialogue task of travel planning.

\begin{figure*}[htb] \centering  
\includegraphics[width=2\columnwidth]{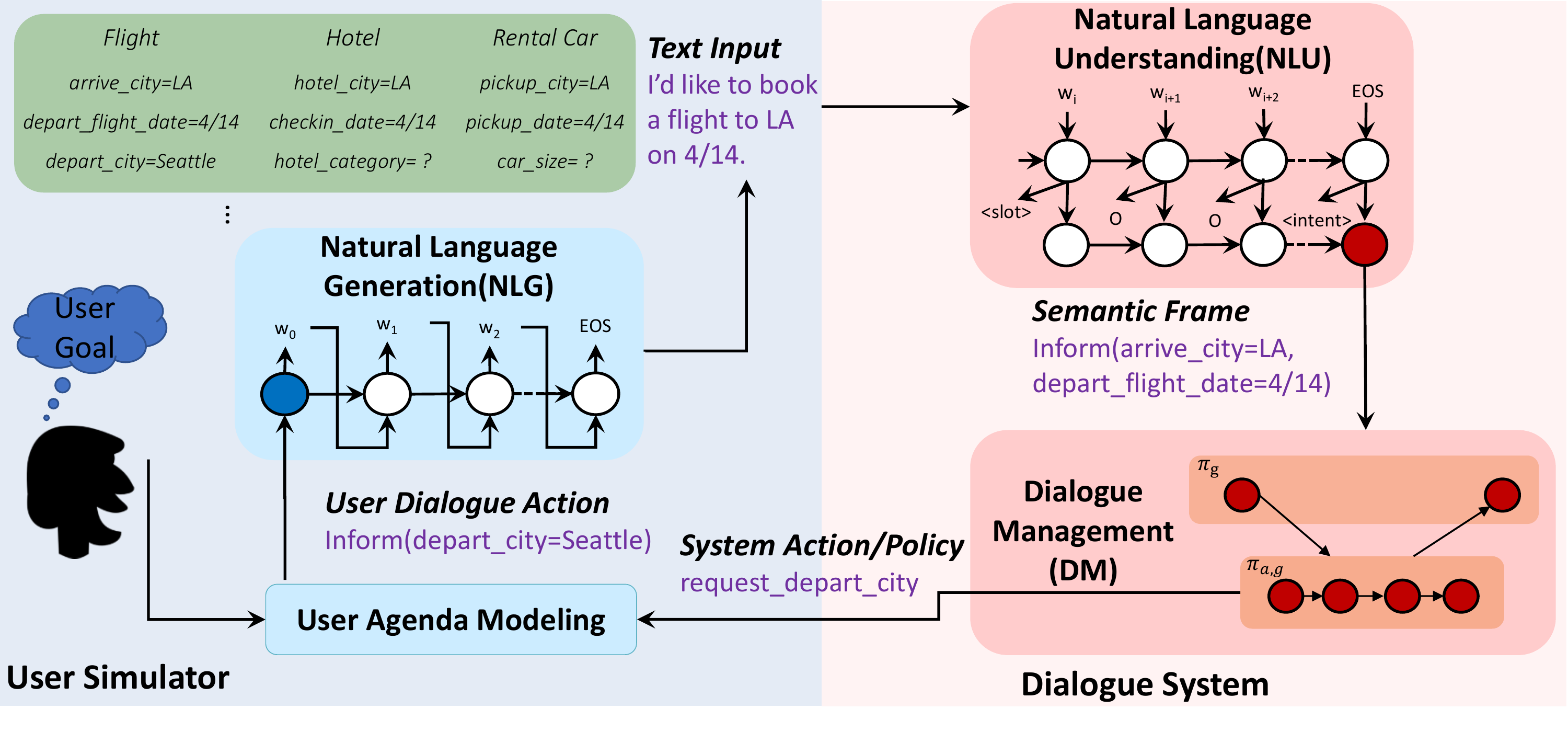}
\caption{Illustration of the Composite Task-Completion dialogue System 
} 
\label{fig:dialog_framework} 
\end{figure*}

\subsection{Dataset}
\label{sec:dataset}

In the study, we made use of a human-human conversation data derived from a publicly available multi-domain dialogue corpus\footnote{https://datasets.maluuba.com/Frames}~\cite{elframes}, which was collected using the Wizard-of-Oz approach. We made a few changes to the schema of the data set for the composite task-completion dialogue setting. Specifically, we added inter-subtask constraints as well as user preferences (soft constraints). The data was mainly used to create simulated users, as will be explained below shortly. 

\subsection{Baseline Agents}
We benchmark the proposed \textit{HRL agent} against three baseline agents: 
\begin{itemize}
\item A \textit{Rule Agent} uses a sophisticated hand-crafted dialogue policy, which requests and informs a hand-picked subset of necessary slots, and then confirms with the user about the reserved tickets.
\item A \textit{Rule+ Agent} requests and informs all the slots in a pre-defined order exhaustedly, and then confirms with the user about the reserved tickets.  The average turn of this agent is longer than that of the \textit{Rule} agent.
\item A \textit{flat RL Agent} is trained with a standard flat deep reinforcement learning method (DQN) which learns a flat dialogue policy using extrinsic rewards only.
\end{itemize}

\subsection{User Simulator}
Training reinforcement learners is challenging because they need an environment to interact with. 
In the dialogue research community, it is common to use simulated users as shown in Figure~\ref{fig:dialog_framework} for this purpose~\cite{DBLP:conf/naacl/SchatzmannTWYY07,DBLP:conf/interspeech/AsriHS16}. In this work, we adapted the publicly-available user simulator, developed by \citet{li2016user}, to the composite task-completion dialogue setting using the human-human conversation data described in Section~\ref{sec:dataset}.\footnote{A detailed description of the user simulator is presented in Appendix A.} During training, the simulator provides the agent with an (extrinsic) reward signal at the end of the dialogue. A dialogue is considered to be successful only when a travel plan is made successfully, and the information provided by the agent satisfies user's constraints. At the end of each dialogue, the agent receives a positive reward of $2*max\_turn$ ($max\_turn=60$ in our experiments) for success, or a negative reward of $-max\_turn$ for failure.  Furthermore, at each turn, the agent receives a reward of $-1$ so that shorter dialogue sessions are encouraged.

\paragraph{User Goal} 
A user goal is represented by a set of slots, indicating the user's request, requirement and preference. For example, an \textit{inform slot}, such as \textsf{dst\_city=``Honolulu''}, indicates a user requirement, and a \textit{request slot}, such as \textsf{price=``?''}, indicates a user asking the agent for the information.

In our experiment, we compiled a list of user goals using the slots collected from the human-human conversation data set described in Section~\ref{sec:dataset}, as follows. We first extracted all the slots that appear in dialogue sessions. If a slot has multiple values, like ``\textsf{or\_city=[San Francisco, San Jose]}'', we consider it as a user preference (soft constraint) which the user may later revise its value to explore different options in the course of the dialogue. If a slot has only one value, we treat it as a user requirement (hard constraint), which is unlikely negotiable. If a slot is with value \textsf{"?"}, we treat it as a user request. We removed those slots from user goals if their values do not exist in our database. The compiled set of user goals contains $759$ entries, each containing slots from at least two subtasks: \emph{book-flight-ticket} and \emph{reserve-hotel}.

\paragraph{User Type} To compare different agents' ability to adapt to user preferences, we also constructed three additional user goal sets, representing three different types of (simulated) users, respectively:
\begin{itemize}
\item \textit{Type A}: All the informed slots in a user goal have a single value.  These users have hard constraints for both the flight and hotel, and have no preference on which subtask to accomplish first. 
\item \textit{Type B}: At least one of informed slots in the \textit{book-flight-ticket} subtask can have multiple values, and the user (simulator) prefers to start with the \textit{book-flight-ticket} subtask. If the user receives ``no ticket available" from the agent during the conversation, she is willing to explore alternative slot values.
\item \textit{Type C}: Similar to \emph{Type B}, at least one of informed slots of the \textit{reserve-hotel} subtask in a user goal can have multiple values. The user prefers to start with the \textit{reserve-hotel} subtask. If the user receives a ``no room available" response from the agent, she is willing to explore alternative slot values.
\end{itemize}


\begin{table*}[htbp]
\bigskip
\begin{center}
\begin{tabular}{ccccccccccc}
\Xhline{2\arrayrulewidth}
\multirow{2}{0.1cm}{} & \multicolumn{3}{c}{Type A} & \multicolumn{3}{c}{Type B} & \multicolumn{3}{c}{Type C} \\ \hline
Agent & Succ. & Turn & Reward & Succ. & Turn & Reward & Succ. & Turn & Reward\\ \hline \hline
Rule & .322 & 46.2 & -24.0 & .240 & 54.2 & -42.9 & .205 & 54.3 & -49.3 \\ 
\textit{Rule+} & \textit{.535} & \textit{82.0} & \textit{-3.7} & \textit{.385} & \textit{110.5} & \textit{-44.95} & \textit{.340} & \textit{108.1} & \textit{-51.85} \\
RL & .437 & 45.6 & -3.3 & .340 & 52.2 & -23.8 & .348 & 49.5 & -21.1 \\
HRL & \textbf{\textcolor{blue}{.632}} & \textbf{\textcolor{blue}{43.0}} & \textbf{\textcolor{blue}{33.2}} & \textbf{\textcolor{blue}{.600}} & \textbf{\textcolor{blue}{44.5}} & \textbf{\textcolor{blue}{26.7}} & \textbf{\textcolor{blue}{.622}} & \textbf{\textcolor{blue}{42.7}} & \textbf{\textcolor{blue}{31.7}} \\
\hline
\end{tabular}
\end{center}
\vspace{-2mm}
\caption{Performance of three agents on different User Types. Tested on 2000 dialogues using the best model during training. Succ.: success rate, Turn: average turns, Reward: average reward. 
}
\vspace{-3mm}
\label{tab:results}
\end{table*}

\begin{figure*}[htb] \centering 
\subfigure[Success Rate of User Type A] { \label{fig:learning_curve_typeA} 
\includegraphics[width=0.66\columnwidth]{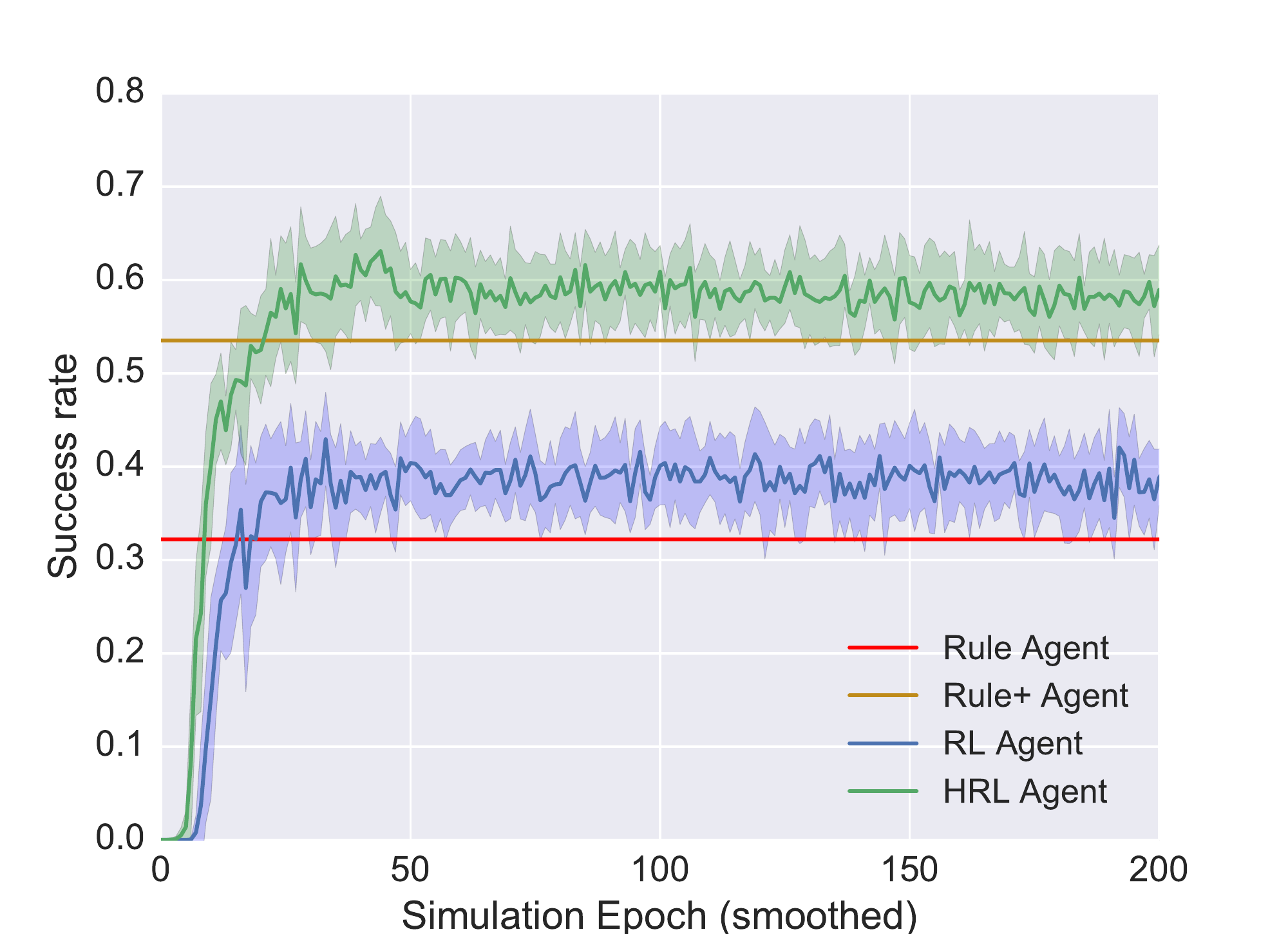}}
\subfigure[Success Rate of User Type B] { \label{fig:learning_curve_typeB} 
\includegraphics[width=0.66\columnwidth]{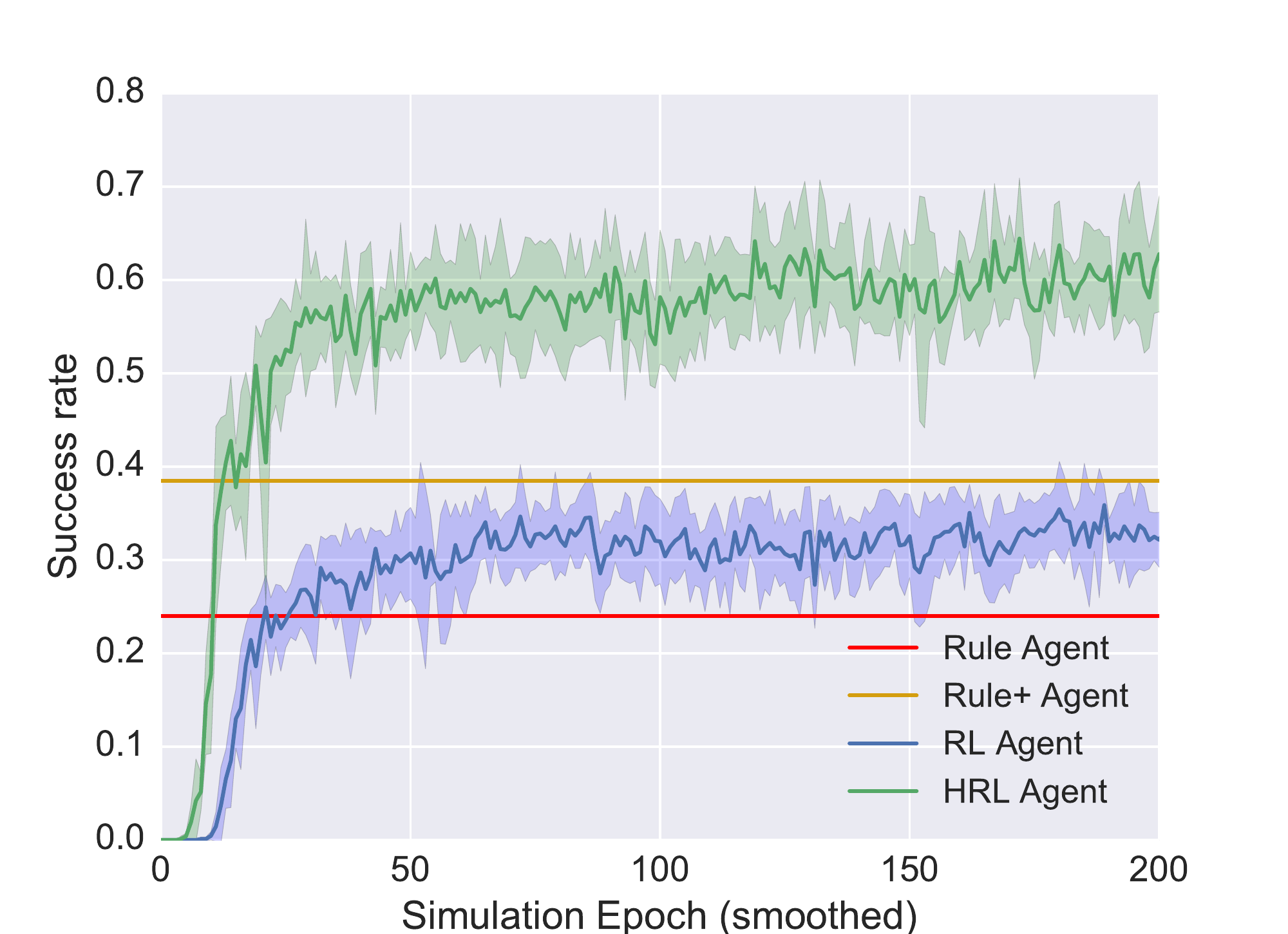}}
\subfigure[Success Rate of User Type C] { \label{fig:learning_curve_typeC} 
\includegraphics[width=0.66\columnwidth]{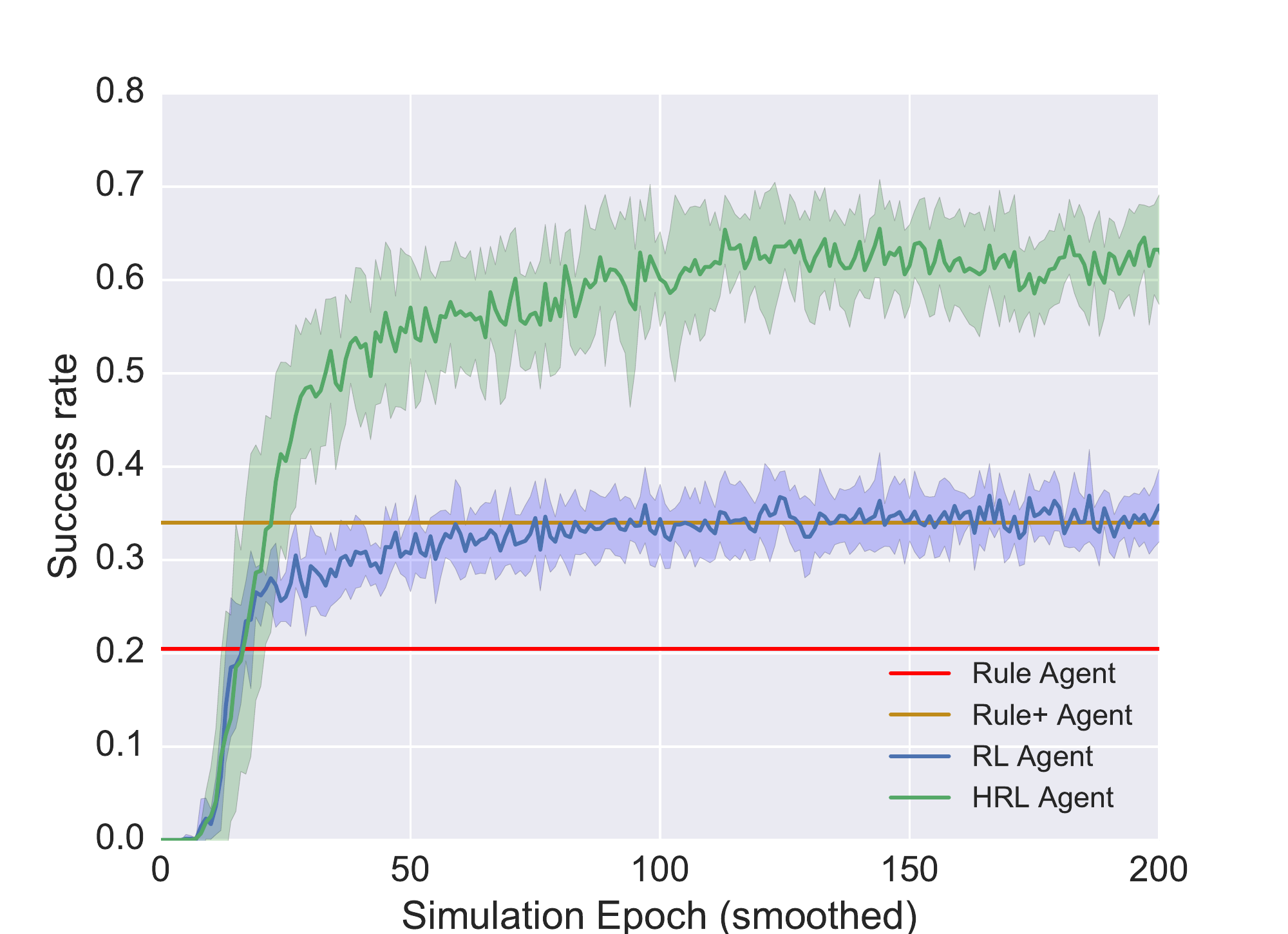}}
\vspace{-2mm}
\caption{Learning curves of dialogue policies for different User Types under simulation} 
\label{fig:learning_curve} 
\vspace{-3mm}
\end{figure*}


\subsection{Implementation}
For the RL agent, we set the size of hidden layer to $80$. For the HRL agent, both top-level and low-level dialogue policies had a hidden layer size of $80$. RMSprop was applied to optimize the parameters. We set batch size to $16$. During training, we used the $\epsilon$-greedy strategy for exploration. For each simulation epoch, we simulated $100$ dialogues and stored these state transition tuples in an experience replay buffer. At the end of each simulation epoch, the model was updated with all the transition tuples in the buffer in a batch manner.

The experience replay strategy is critical to the success of deep reinforcement learning. In our experiments, at the beginning, we used a rule-based agent to run $N$ ($N=100$) dialogues to populate the experience replay buffer, which was an implicit way of imitation learning to initialize the RL agent. Then, the RL agent accumulated all the state transition tuples and flushes the replay buffer only when the current RL agent reached a success rate threshold no worse than that of the \textit{Rule} agent.

This strategy was motivated by the following observation. The initial performance of an RL agent was often not strong enough to result in dialogue sessions with a reasonable success rate.  With such data, it was easy for the agent to learn the locally optimal policy that ``failed fast''; that is, the policy would finish the dialogue immediately, so that the agent could suffer the least amount of per-turn penalty.  Therefore, we provided some rule-based examples that succeeded reasonably often, and did not flush the buffer until the performance of the RL agent reached an acceptable level. Generally, one can set the threshold to be the success rate of the \textit{Rule} agent. To make a fair comparison, for the same type of users, we used the same \textit{Rule} agent to initialize both the RL agent and the HRL agent.

	\begin{table*}[bt!]
		\footnotesize
		\centering
		\caption{Sample dialogue by RL and HRL agents with real user: Left column shows the dialogue with the RL agent; Right column shows the dialogue with the HRL agent; bolded slots are the joint constraints between two subtasks.}
		\begin{tabular}[t]{ll}
			\hline
			\multicolumn{2}{c}{User Goal} \\
			\begin{tabular}[ct]{@{}l@{}}
							\textit{reserve-hotel} subtask:\\
				\{\\
				\-\hspace{1mm} ``request\_slots": \{	 \-\hspace{13mm}     ``inform\_slots": \{\\
				\-\hspace{4mm}    ``hotel\_price": ``?"	\-\hspace{12mm}    \textbf{``hotel\_city": ``Cancun"},\\
				\-\hspace{4mm}    ``hotel\_date\_checkout": ``?"	\-\hspace{0mm}    \textbf{``hotel\_numberofpeople": ``3"},	\\
				\-\hspace{4mm}    ``hotel\_name": ``?"	\-\hspace{11mm}    \textbf{``hotel\_date\_checkin": ``09/20"}\\
				\-\hspace{1mm}  \},		\-\hspace{34mm}  \}\\
				\}
                \\
                \\
				\-\hspace{3mm}

			\end{tabular}  & \begin{tabular}[ct]{@{}l@{}}
				\textit{book-flight-ticket} subtask:\\
				\{\\
				\-\hspace{1mm} ``request\_slots": \{ \-\hspace{8mm}     ``inform\_slots": \{\\
				\-\hspace{4mm}    ``price": ``?"	\-\hspace{15.5mm}    ``or\_city": ``Campinas",	\\
				\-\hspace{4mm}    ``return\_time\_dep": ``?"	\-\hspace{2mm}   \textbf{``dst\_city": ``Cancun"},\\
				\-\hspace{4mm}    ``return\_date\_dep": ``?"	\-\hspace{2.5mm}    \textbf{``numberofpeople": ``3"},\\
				\-\hspace{4mm}    ``depart\_time\_dep": ``?"				\-\hspace{2mm}    \textbf{``depart\_date\": "9/20"}	\\
				\-\hspace{4mm}    ``seat": ``?"	\-\hspace{16mm}  \}\\
				\-\hspace{1mm}  \}	\\
				\}\\
			\end{tabular} \\
			\hline
			\multicolumn{1}{c|}{\textbf{RL agent}} & \multicolumn{1}{c}{\textbf{HRL agent}} \\
			\multicolumn{1}{l|}{
				\textit{Agent}: Welcome to our travel agent. May I know your} & \textit{Agent}: Welcome to our travel agent. May I know your\\
			\multicolumn{1}{l|}{
				original city, destination city and departure date?} & original city, destination city and departure date? \\
			\multicolumn{1}{l|}{
				\textit{User}: I depart from Campinas to Cancun and prefer to} & \textit{User}: I depart from Campinas to Cancun and prefer to\\
			\multicolumn{1}{l|}{
				leave on 09-20.} & leave on 09-20. \\
			\multicolumn{1}{l|}{
				\textit{Agent}: There is a return flight on 9/26 and hotel checkout date} & \textit{Agent}: Only Business class seat is available. \\
			\multicolumn{1}{l|}{
				is on 9/26 too.} &  \\
			\multicolumn{1}{l|}{
				\textit{User}: My wife and son will travel with me, three people in total.} & \textit{User}: What is the depart time? \\
			\multicolumn{1}{l|}{
				\textit{Agent}: \textbox{blue}{There is a hotel called Hotel Tropic available for you.}} & \textit{Agent}:	Departure time is at 10:00AM. \\
			\multicolumn{1}{l|}{
				\textit{User}: What depart time is available?} & \textit{User}: My wife and son will travel with me, \\
			\multicolumn{1}{l|}{
				 } & three people in total. \\
			\multicolumn{1}{l|}{
				\textit{Agent}: Departure time is at 10:00AM.} & \textit{Agent}: There is one return flight departed at 4:00PM. \\
			\multicolumn{1}{l|}{
				\textit{User}: What is the departure time of return flight?} & \textit{User}: What is the date of the return flight? \\
			\multicolumn{1}{l|}{
				\textit{Agent}: It is 4:00PM.} & \textit{Agent}: It is on 9/26 and hotel checkout date  is also 9/26.\\
			\multicolumn{1}{l|}{
				\textit{User}: Which seat is available?} & \textit{User}:	What's the price for the flight? \\
			\multicolumn{1}{l|}{
				\textit{Agent}: Only Business class available.} & \textit{Agent}: Total is 1399 dollars for flight. \\
			\multicolumn{1}{l|}{
				\textit{User}: What's the price for the flight?} & \textit{User}: Which hotel is available? \\
			\multicolumn{1}{l|}{
				\textit{Agent}: Total is 1399 dollars for flight.} & \textit{Agent}: A hotel called Hotel Tropic is available for you. \\
			\multicolumn{1}{l|}{
				\textit{User}: What's the price for the hotel?} & \textit{User}: What's the price for the hotel? \\
			\multicolumn{1}{l|}{
				\textit{Agent}: It is 1091 dollars for hotel.} & \textit{Agent}: It is 1091 dollars for hotel. \\
			\multicolumn{1}{l|}{
				\textit{User}: Could you please help me book the tickets?} & \textit{User}: Could you please help me book the tickets? \\
			\multicolumn{1}{l|}{
				\textit{Agent}: Departure flight on 9/20 and hotel checkin date} & \textit{Agent}: Okay. Tickets are issued. \\
			\multicolumn{1}{l|}{
				on 9/20 are available.}  & \\
			\multicolumn{1}{l|}{
				\textit{User}: Could you please help me book the tickets?} & \textit{User}: Thanks! \\
			\multicolumn{1}{l|}{
				\textit{Agent}: Okay. Tickets are issued.} & \textit{Agent}: Thanks! \\
			\multicolumn{1}{l|}{
				\textit{User}: Thanks!} & \\
			\multicolumn{1}{l|}{
				\textit{Agent}: Thanks!} & \\
			\hline
			\multicolumn{1}{l|}{
			\textbf{Success}: but the agent does not behave naturally \textbox{blue}{where} is an } & \multicolumn{1}{c}{\textbf{Success}}  \rule{0pt}{1.1\normalbaselineskip} \\
			\multicolumn{1}{l|}{inter-subtask transition from \textit{book-flight-ticket} to \textit{reserve-hotel}.} &  \\ \hline
		\end{tabular}
		\label{tab:sample_dialogue}
        \vspace{-2mm}
	\end{table*}

\begin{figure}[htb]
\centering
\includegraphics[width=1\linewidth]{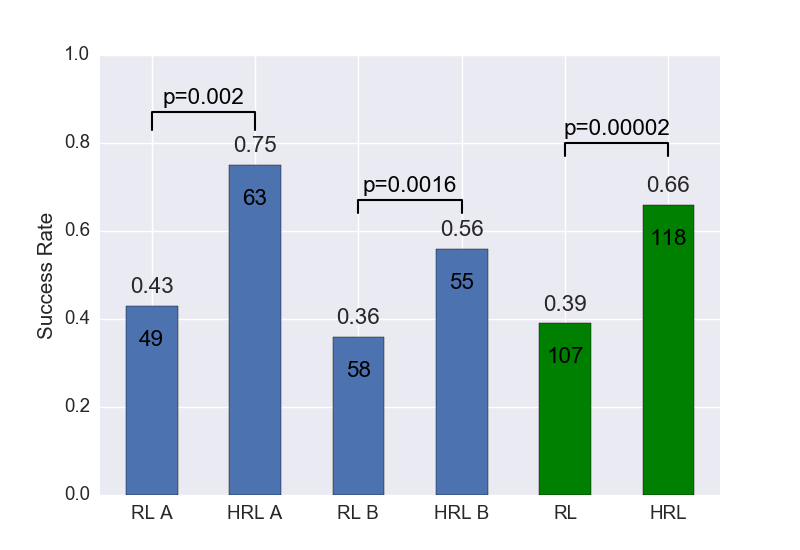}
\vspace{-6mm}
\caption{Performance of HRL agent versus RL agent tested with real users: success rate, number of tested dialogues and p-values are indicated on each bar; the rightmost green ones are for total (difference in mean is significant with $p <$ 0.01).}
\label{fig:user_success_rate}
\vspace{-3mm}
\end{figure}
\begin{figure}[htb]
\centering
\includegraphics[width=1\linewidth]{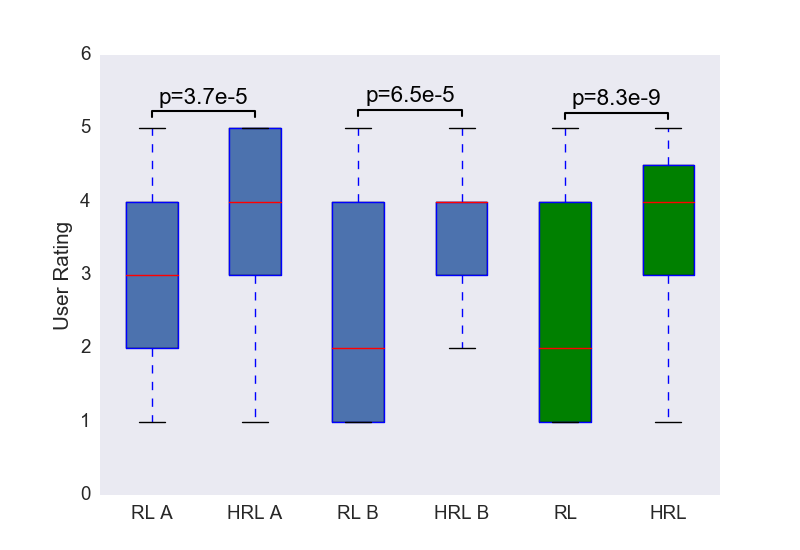}
\vspace{-6mm}
\caption{Distribution of user ratings for HRL agent versus RL agent, and total.}
\label{fig:user_rating}
\vspace{-2mm}
\end{figure}

\subsection{Simulated User Evaluation}

On the composite task-completion dialogue task, we compared the HRL agent with the baseline agents in terms of three metrics: success rate\footnote{Success rate is the fraction of dialogues where the tasks are successfully accomplished within the maximum turns. 
}, average rewards, and the average number of turns per dialogue session.


Figure~\ref{fig:learning_curve} shows the learning curves of all four agents trained on different types of users. Each learning curve was averaged over $10$ runs. Table~\ref{tab:results} shows the performance on test data. For all types of users, the HRL-based agent yielded more robust dialogue policies outperforming the hand-crafted rule-based agents and flat RL-based agent measured on success rate.  It also needed fewer turns per dialogue session to accomplish a task than the rule-based agents and flat RL agent.
The results across all three types of simulated users suggest the following conclusions.

First, he HRL agent significantly outperformed the RL agent. This, to a large degree, was attributed to the use of the hierarchical structure of the proposed agent. Specifically, the top-level dialogue policy selected a subtask for the agent to focus on, one at a time, thus dividing a complex task into a sequence of simpler subtasks. The selected subtasks, combined with the use of intrinsic rewards, alleviated the sparse reward and long-horizon issues, and helped the agent explore more efficiently in the state-action space. As a result, as shown in Figure~\ref{fig:learning_curve} and Table~\ref{tab:results}, the performance of the HRL agent on types B and C users (who may need to go back to revise some slots during the dialogue) does not drop much compared to type A users, despite the increased search space in the former. Additionally, we observed a large drop in the performance of the RL Agent due to the increased complexity of the task, which required more dialogue turns and posed a challenge for temporal credit assignment.

Second, the HRL agent learned much faster than the RL agent. The HRL agent could reach the same level of performance with a smaller number of simulation examples than the RL agent, 
demonstrating that the hierarchical dialogue policies were more sample-efficient than flat RL policy and could significantly reduce the sample complexity on complex tasks.

Finally, we also found that the \textit{Rule}+ and \textit{flat} RL agents had comparable success rates, as shown in Figure~\ref{fig:learning_curve}. However, a closer look at the correlation between success rate and the average number of turns in Table~\ref{tab:results} suggests that the \textit{Rule}+ agent required more turns which adversely affects its success, whilst the \textit{flat} RL agent achieves similar success with much less number of turns in all the user types. It suffices to say that our hierarchical RL agent outperforms all in terms of success rate as depicted in Figure~\ref{fig:learning_curve}.

\subsection{Human Evaluation}

We further evaluated the agents, which were trained on simulated users, against real human users, recruited from the authors' affiliation. We conducted the study using the HRL and RL agents, each tested against two types of users: \emph{Type A} users who had no preference for subtask, and \emph{Type B} users who preferred to complete the \textit{book-flight-ticket} subtask first. Note that \emph{Type C} users were symmetric to Type B ones, so were not included in the study.  We compared two (agent, user type) pairs: \{RL A, HRL A\} and \{RL B, HRL B\}; in other words, four agents were trained against their specific user types. In each dialogue session, one of the agents was randomly picked to converse with a user. 
The user was presented with a user goal sampled from our corpus, and was instructed to converse with the agent to complete the task. If one of the slots in the goal had multiple values, the user had multiple choices for this slot and might revise the slot value when the agent replied with a message like ``No ticket is available" during the conversation. At the end of each session, the user was asked to give a rating on a scale from $1$ to $5$ based on the naturalness and coherence of the dialogue. ($1$ is the worst rating, and $5$ the best). We collected a total of $225$ dialogue sessions from $12$ human users.

Figure~\ref{fig:user_success_rate} presents the performance of these agents against real users in terms of success rate. Figure~\ref{fig:user_rating} shows the comparison in user rating. For all the cases, the HRL agent was consistently better than the RL agent in terms of success rate and user rating. Table~\ref{tab:sample_dialogue} shows a sample dialogue session. We see that the HRL agent produced a more coherent conversation, as it switched among subtasks much less frequently than the \textit{flat} RL agent.


\section{Discussion and Conclusions}

This paper considers composite task-completion dialogues, where a set of subtasks need to be fulfilled collectively for the entire dialogue to be successful. We formulate the policy learning problem using the options framework, and take a hierarchical deep RL approach to optimizing the policy. Our experiments, both on simulated and real users, show that the hierarchical RL agent significantly outperforms a flat RL agent and rule-based agents. The hierarchical structure of the agent also improves the coherence of the dialogue flow.

The promising results suggest several directions for future research. First, the hierarchical RL approach demonstrates strong adaptation ability to tailor the dialogue policy to different types of users. This motivates us to systematically investigate its use for dialogue \textit{personalization}. Second, our hierarchical RL agent is implemented using a two-level dialogue policy. But more complex tasks might require multiple levels of hierarchy. Thus, it is valuable to extend our approach to handle such deep hierarchies, where a subtask can invoke another subtask and so on, taking full advantage of the options framework. Finally, designing task hierarchies requires substantial domain knowledge and is time-consuming. This challenge calls for future work on automatic learning of hierarchies for complex dialogue tasks.

\section*{Acknowledgments}
Baolin Peng was in part supported by General Research Fund of Hong Kong (14232816). We would like thank anonymous reviewers for their insightful comments.

\bibliography{emnlp2017}
\bibliographystyle{emnlp_natbib}

\newpage
\appendix

\section{User Simulator}
\label{app:appendix_us}
\paragraph{User Goal} In the task-completion dialogue setting, the first step of user simulator is to generate a feasible user goal. Generally, a user goal is defined with two types of slots: \emph{request} slots that user does not know the value and expects the agent to provide it through the conversation; \emph{inform} slots is slot-value pairs that user know in the mind, serving as \emph{soft}/\emph{hard} constraints in the dialog; slots that have multiple values are termed as \emph{soft} constraints, which means user has preference, and user might change its value when there is no result returned from the agent based on the current values; otherwise, slots that have with only one value serve as \emph{hard} constraint. Table~\ref{tab:user_goal} shows an example of a user goal in the composite task-completion dialogue.

\begin{table}[htbp]
\small
\centering
\begin{tabular}{|c|c|c|}
\hline 
& \textit{book-flight-ticket} & \textit{reserve-hotel}\\ 
\hline
\multirow{5}{*}{\begin{sideways}inform\end{sideways}} & \textbf{dst\_city}=LA & \textbf{hotel\_city}=LA \\
& \textbf{numberofpeople}=2 & \textbf{hotel\_numberofpeople}=2\\ 
& \textbf{depart\_date\_dep}=09-04 & \textbf{hotel\_date\_checkin}=09-04\\
& or\_city=Toronto &\\
& seat=economy &\\
\hline
\multirow{4}{*}{\begin{sideways}request\end{sideways}}& price=?&hotel\_price=?\\
& return\_time\_dep=?&hotel\_date\_checkout=?\\
& return\_date\_dep=?&hotel\_name=?\\
& depart\_time\_dep=?&\\
\hline
\end{tabular}
\caption{An example of user goal}
\label{tab:user_goal}
\end{table}

\paragraph{First User Act} 
This work focuses on user-initiated dialogues, so we randomly generate a user action as the first turn (a user turn). To make the first user-act more reasonable, we add some constraints in the generation process. For example, the first user turn can be inform or request turn; it has at least two informable slots, if the user knows the original and destination cities, \emph{or\_city} and \emph{dst\_city} will appear in the first user turn etc.; If the intent of first turn is request, it will contain one requestable slot.

During the course of a dialogue, the user simulator maintains a compact stack-like representation named as \emph{user agenda}~\cite{schatzmann2009hidden}, where the user state $s_u$ is factored into an agenda $A$ and a goal $G$, which consists of constraints $C$ and request $R$. At each time-step $t$, the user simulator will generate the next user action $a_{u,t}$ based on the its current status $s_{u,t}$ and the last agent action $a_{m,t-1}$, and then update the current status $s'_{u,t}$. Here, when training or testing a policy without natural language understanding (NLU) module, an error model~\cite{li2017investigation} 
is introduced to simulate the noise from the NLU component, and noisy communication between the user and agent.

\section{Algorithms}
\label{app:appendix_algo}
Algorithm~\ref{algo:hdqn} outlines the full procedure for training hierarchical dialogue policies in this composite task-completion dialogue system.

\begin{algorithm*}[!bt]
\caption{Learning algorithm for HRL agent in composite task-completion dialogue}
\begin{algorithmic}[1]
\STATE Initialize experience replay buffer $\mathcal{D}_1$ for meta-controller and $\mathcal{D}_2$ for controller.
\STATE Initialize $Q_1$ and $Q_2$ network with random weights.
\STATE Initialize dialogue simulator and load knowledge base.
\FOR {$episode$=1:N}
\STATE Restart dialogue simulator and get state description $s$
\WHILE{s is not terminal}
\STATE $extrinsic\_reward := 0$
\STATE $s_0 := s$
\STATE select a subtask $g$ based on probability distribution $\pi(g|s)$ and exploration probability $\epsilon_g$
\WHILE {s is not terminal and subtask g is not achieved}
\STATE select an action $a$ based on the distribution $\pi(a|s,g)$ and exploration probability $\epsilon_c$
\STATE Execute action $a$, obtain next state description $s'$, perceive extrinsic reward $r^e$ from environment
\STATE Obtain intrinsic reward  $r^i$ from internal critic
\STATE Sample random minibatch of transitions from $\mathcal{D}_1$
\STATE	$y = \begin{cases} r^i &\mbox{if} s' \mbox{ is terminal} \\ r^i + \gamma * max_{a'} Q_1(\{s',g\},a';\theta_1) & \mbox{oterwise}\end{cases}$
\STATE Perform gradient descent on loss $\mathcal{L}(\theta_1)$ according to equation \ref{eq:g1}
\STATE Store transition(\{$s$,$g$\},a,$r^i$,\{$s'$,$g$\}) in $\mathcal{D}_1$
\STATE Sample random minibatch of transitions from $\mathcal{D}_2$
\STATE	$y = \begin{cases} r^e &\mbox{if} s' \mbox{ is terminal} \\ r^e + \gamma * max_{a'} Q_2(s',g',a';\theta_2) & \mbox{oterwise}\end{cases}$
\STATE Perform gradient descent on loss $\mathcal{L}(\theta_2)$  according to equation \ref{eq:g2}
\STATE $extrinsic\_reward$ += $r^e$
\STATE $s = s'$
\ENDWHILE
\STATE Store transition ($s_0$, $g$, $extrinsic\_reward$, $s'$) in $\mathcal{D}_2$
\ENDWHILE
\ENDFOR
\end{algorithmic}
\label{algo:hdqn}
\end{algorithm*}
    
\end{document}